%% file: main.tex
\pdfoutput=1

\documentclass[11pt]{article}

\usepackage[]{acl}

\usepackage{times}
\usepackage{latexsym}

\usepackage{pifont}
\usepackage{adjustbox}
\usepackage{multirow}

\usepackage{enumitem}
\setlist{itemsep=1.5pt,parsep=1.5pt}             

\usepackage[T1]{fontenc}

\usepackage[utf8]{inputenc}

\usepackage{microtype}

\usepackage{amsmath}

\usepackage{graphicx}
\usepackage{booktabs}
\usepackage{comment}

\usepackage{xcolor, soul}
\sethlcolor{green}

\usepackage{color, colortbl}
\definecolor{LightCyan}{rgb}{0.88,1,1}
\definecolor{LightGreen}{rgb}{0.72,0.91,0.72}

\usepackage{subcaption,booktabs}

\title{PruMUX: Augmenting Data Multiplexing with Model Compression}

\author{Yushan Su \quad Vishvak Murahari \quad Karthik Narasimhan \quad Kai Li \\
        Princeton University\\
        Princeton, NJ, USA\\
        \texttt{\{yushans, murahari, karthikn, li\}@princeton.edu }}

\newcommand{\prumux}{PruMUX}
\newcommand{\autoprumux}{Auto-PruMUX}
\newcommand{\DataMUX}{DataMUX}

\begin{document}
\maketitle
\begin{abstract}
As language models increase in size by the day, methods for efficient inference are critical to leveraging their capabilities for various applications. Prior work has investigated techniques like model pruning, knowledge distillation, and data multiplexing to increase model throughput without sacrificing accuracy. In this paper, we combine two such methods -- structured pruning and data multiplexing -- to compound the speedup gains obtained by either method. Our approach, PruMUX, obtains up to 7.5-29.5X throughput improvement over BERT-base model with accuracy threshold from 80\% to 74\%. We further study various combinations of parameters (such as sparsity and multiplexing factor) in the two techniques to provide a comprehensive analysis of the tradeoff between accuracy and throughput in the resulting models. We then propose Auto-PruMUX, a meta-level model that can predict the high-performance parameters for pruning and multiplexing given a desired accuracy loss budget, providing a practical method to leverage the combination effectively.\footnote{Our code is available at \url{https://github.com/yushansu/PruMUX}}

\end{abstract}

\input{intro}

\input{background.tex}
\input{prumux}

\input{prumuxauto.tex}
\input{related_work}

\input{conclusion}

\input{limitations}

\bibliography{anthology,custom}
\bibliographystyle{acl_natbib}

\input{appendix}

\end{document}

%% file: intro.tex
\section{Introduction}

Large language models (LLMs) have achieved state-of-the-art performance across various NLP tasks and resulted in impressive user-facing demonstrations such as ChatGPT.\footnote{\url{https://chat.openai.com/}} However, their large size necessitates the use of enormous amounts of compute and memory at inference time, which limits their widespread use. 

Two types of techniques have been explored to reduce the cost of model inference.  The first is model compression including network pruning~\cite{lecun89, han15, frankle19}, quantization~\cite{han16}, knowledge distillation~\cite{hinton2015distilling}, combinations of multiple methods~\cite{xia-etal-2022-structured}.  
The second is recently proposed data multiplexing~\cite{murahari2023mux}, which multiplexes multiple inputs into a single input for model inference.

While both types of methods leverage the over-parameterization effect~\cite{allen2019convergence, radhakrishnan2020overparameterized} in modern deep neural networks to improve the throughput-to-compute cost ratio, the manner in which they do so is different. Model compression aims at reducing the number of parameters in the model, hence reducing the overall compute cost (denominator) to improve the ratio.  Data multiplexing, on the other hand, compresses multiple inputs into one to improve throughput (numerator) while keeping the model size fixed.  This observation naturally leads us to hypothesize that the two types of methods could be complementary and can be combined for maximal gain in the throughput-to-compute cost ratio.  

\begin{figure}[t]
\centering
\includegraphics[width=0.45\textwidth]{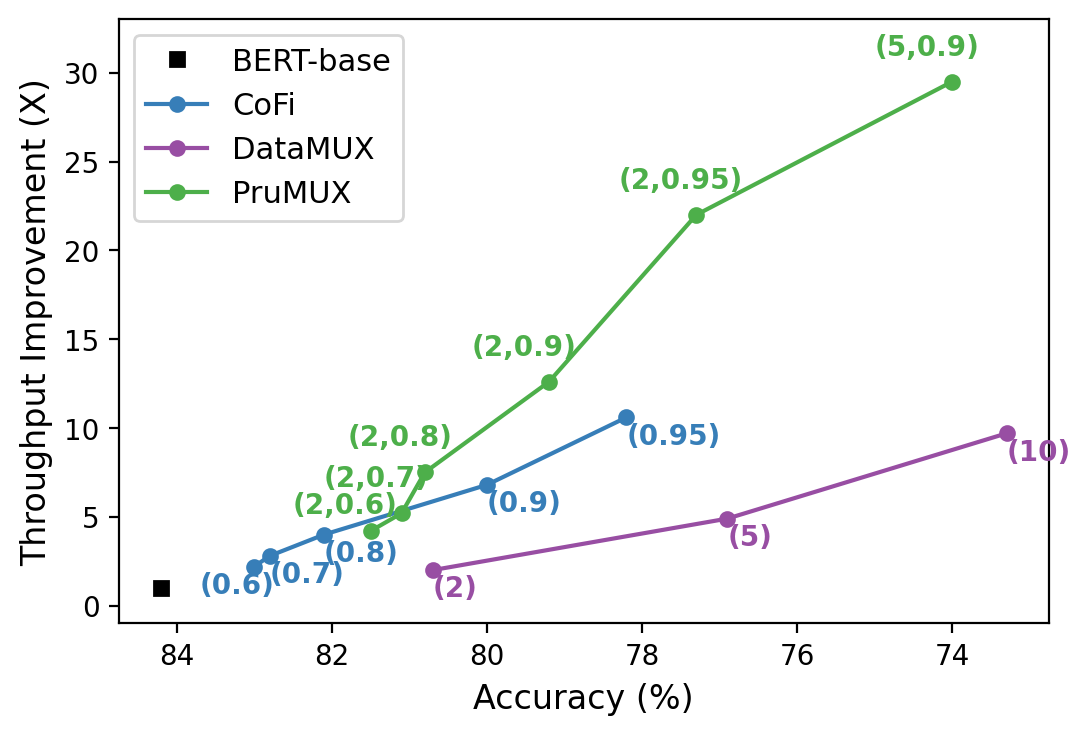}
\caption{\label{fig:intro} Throughput improvements ($\times$) of CoFi, \DataMUX{}, and PruMUX over the BERT-base model~\cite{devlin2018bert} on the MNLI task~\cite{williams2017broad}. The sparsity for a CoFi's data point is labeled as $s$.  The width of multiplexing for a DataMUX's data point is labeled as $N$.  The parameter pair for a PruMUX's data point is labeled as ($N$, $s$).
}
\label{fig:teaser}
\end{figure}

There are two challenges to this hypothesis.  The first is that both model compression and data multiplexing aim at trading a small accuracy loss for large throughput improvement.  Intuitively, the combination may incur an accuracy loss larger than either method and it is not clear how they interact with each other when combining them together.  A research question is how to combine the two methods such that the combination achieves better throughput than each type of method individually, given any accuracy loss budget or accuracy threshold.

The second challenge is to efficiently find the best parameters pair ($N,s$) where $N$ is the width of the data multiplexing and $s$ is the sparsity of the model compression method.  Training and testing with each parameter combination is costly and time-consuming.  
A research question is how to automatically predict and find top parameters based on the model's performance on one set of parameters.

To address the first research question, we present \prumux{}, a combination of model compression and data multiplexing.  Our method is simple and consists of three phases -- multiplexed model pre-training, task-specific fine-tuning and task-specific model compression.  In our implementation, we make use of CoFi~\cite{xia-etal-2022-structured}, a state-of-the-art model compression method that includes intermediate knowledge distillation steps that help minimize accuracy hits, and DataMUX~\cite{murahari2023mux}, which performs vector-based input multiplexing over instances. 

Our results over four datasets (MNLI, QNLI, QQP and SST-2) demonstrate that \prumux{} achieves significantly higher throughput over CoFi and DataMUX individually
for a large range of accuracy thresholds.
As an example, Figure~\ref{fig:teaser} shows the throughput improvements over the BERT-base model on task MNLI,   providing a more optimal Pareto frontier in the tradeoff between accuracy and throughput.  

To address the second research question, we propose \autoprumux{}, a meta-model to automatically predict and find the high-performance parameter combinations for a desired accuracy loss budget on a task based on the model's performance on one set of parameters without running additional experiments.
We use interpolation and estimation models over a set of data points to predict the accuracy and throughput of a \prumux{} model based on sparsity and multiplexing factor. We show promise in modeling the tradeoffs accurately and \autoprumux{} can find high-performance combinations of known parameters as well as unknown parameters,  
providing a practical method for choosing a high-performance \prumux{} model for a downstream task.

Our key insight for why \prumux{} can achieve better throughput than model compression and data multiplexing individually is that they improve the throughput of a model in two different dimensions: reducing the latency of an inference and compressing multiple inferences.  In addition,  both methods lead to non-linear drops in model accuracy at some points. \prumux{} can achieve high throughput while avoiding each method's limitations.

%% file: background.tex
\section{Background}

\subsection{CoFi Pruning}
CoFi is a state-of-the-art model compression method~\cite{xia-etal-2022-structured} that uses distillation and structured pruning to jointly prune a Transformer network~\cite{devlin2018bert}.  Its key idea is to  distill the knowledge from the base model into the pruned model during training.  A layer-wise distillation approach is used to guide the pruning from the teacher model, i.e., dense model, to the student model, i.e., pruned model, with a loss defined as:
\begin{equation*}
    L_{layer} = \sum_{i \in \tau}MSE(W_{layer}\textbf{H}_s^{m(i)}, \textbf{H}_t^i)
\end{equation*}

where $\textbf{H}_s^{m(i)}$ and $\textbf{H}_t^i$ are hidden representations of the $m(i)$th feed-forward layer of the student model and $i$th feed-forward layer of the teacher model. $i$ is the teacher model's closest layer to the layer $m(i)$ of the student model. $W_{layer}$ is a linear transformation matrix, initialized as an identity matrix.

CoFi prunes both coarse-grained and fine-grained units of the distilled network. 
The coarse-grained units include multi-head attention layers, fully-connected layers, and attention heads. The fine-grained units include hidden dimensions and intermediate dimensions of the Transformer model. 
Different masks are used for different pruning units and are learned via $\ell_0$  regularization during training. The units with mask variables smaller than a threshold are pruned away before inference.

\subsection{DataMUX}\label{sec:DataMUX}

Data multiplexing (DataMUX) is a recently proposed method~\cite{murahari2022datamux, murahari2023mux} to compress multiple inputs into a single ``mixed'' representation of the same size as a single input to a network, in order to improve inference throughput. 
DataMUX introduces multiplexing layers, which multiplex different sequences into a single sequence of representations, i.e., multiplexed representations, and demultiplexing layers, which demultiplex/decompress the multiplexed representations. The multiplexed layer first compresses multiple input sequences into a single sequence of representations. These representations are then processed by a Transformer model and the resulting representations are then disentangled into independent representations by the demultiplexer layer. These representations are then used to make predictions. DataMUX, therefore, leads to a many-fold increase in inference throughput as just a single pass through the large Transformer model. 

The multiplexing layer is defined as 
\begin{equation*}
    \textbf{x}^{1:N} = \Phi(\textbf{x}^1, ... \textbf{x}^N) = \frac{1}{N}\sum_{i=1}^{N}\phi^i(\textbf{x}^i)
\end{equation*}
where $\textbf{x}$ is the input sequence, $\phi^i, i \in [1,...N]$, is the Hadamard product with a fixed Gaussian random vector and $N$ is the number of input sequences that get multiplexed. The multiplexed representations, $\textbf{x}^{1:N}$, are then processed by the Transformer model to generate hidden multiplexed representations, $\textbf{h}^{1:N}$.

The demultiplexer layer, in order to disentangle the hidden multiplexed representation, $\textbf{h}^{1:N}$, into independent representations, learns N parameterized demultiplexing functions, $\psi^i$. The independent representations, $\textbf{h}^i $, are then used to make predictions.
\begin{equation*}
    \textbf{h}^i = \psi^i(\textbf{h}^{1:N}) \quad \forall i \in 1,2,...N
\end{equation*}

\subsection{Observations}
Both model compression and data multiplexing aim at trading small accuracy losses for large inference throughput improvements. When CoFi prunes a Transformer at relatively low sparsities, its accuracy loss is minimal and throughput improvement is significant, but  at 95\% sparsity, its accuracy loss becomes relatively significant~\cite{xia-etal-2022-structured}. DataMUX also shares this nonlinear property, as shown in Figure~\ref{fig:intro}. In other words, the trade-off of each method is good only up to a certain point.  

The two methods improve the throughput of a model in two dimensions. CoFi reduces the latency of an inference, whereas DataMUX compresses multiple inferences into one.  A natural question is whether combining the two methods can achieve higher throughput with a smaller accuracy loss than each method individually.

%% file: prumux.tex
\section{PruMUX}

\begin{figure}[h]
\centering
\includegraphics[width=0.45\textwidth]{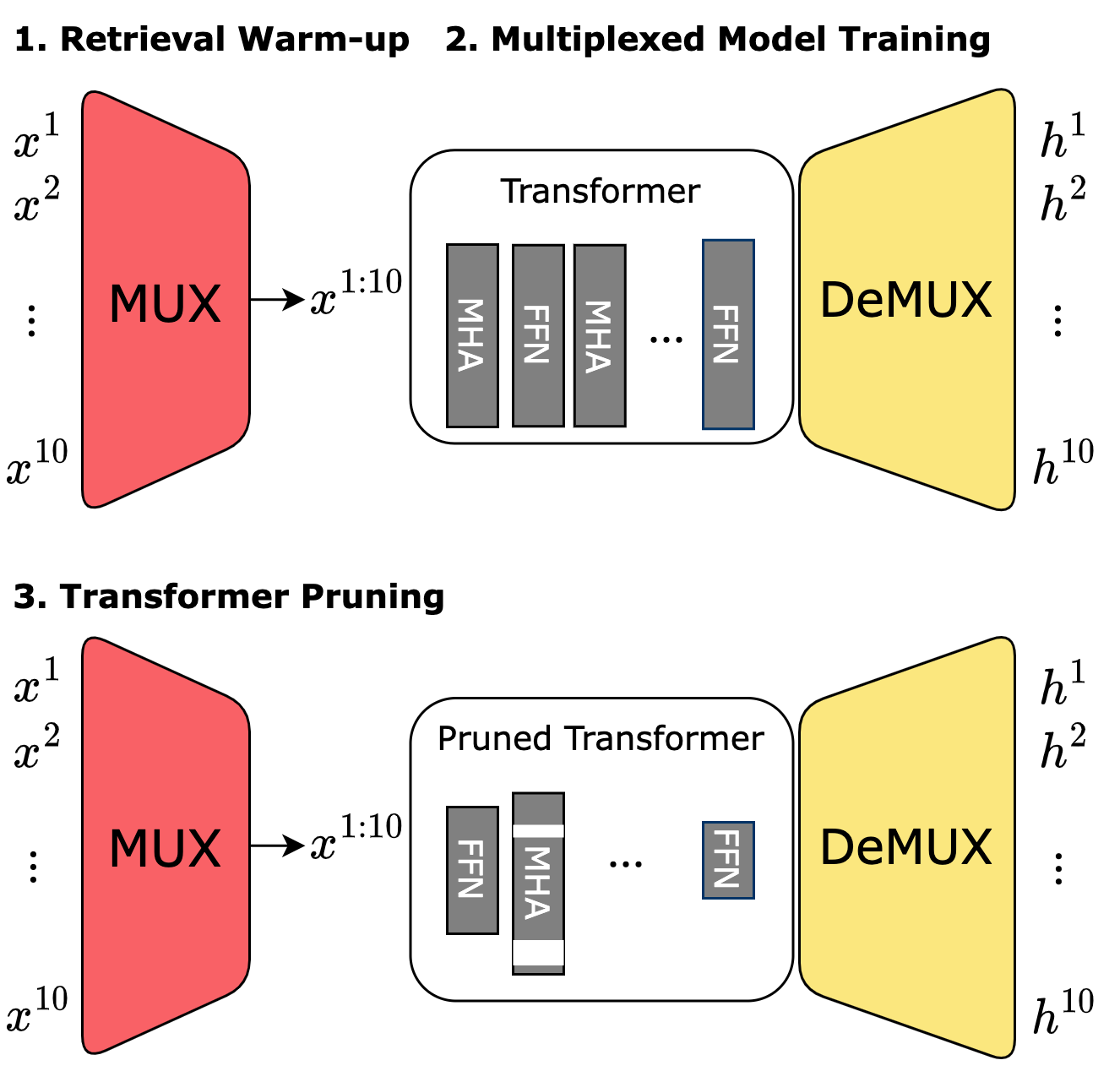}
\caption{\label{fig:overview} 
Illustration of PruMUX showing a multiplexer, sparse Transformer, and a demultiplexer, with multiplexing width of 10, where 10 input sequences are mixed into 1 input sequence. The multiplexed Transformer model is pruned to reduce inference time. The training for PruMUX consists of three steps including retrieval warm-up, multiplexed model training, and Transformer pruning. }
\end{figure}

Our key motivational question is the following: \textit{given an accuracy loss budget, can the combination of model compression and data multiplexing achieve better throughput than each method individually?}
In this section, we first present PruMUX, a method to combine the two methods, and then show that PruMUX achieves substantially better throughput than each method alone for various accuracy thresholds in our experimental results.

\subsection{Method}

PruMUX is a method to convert any Transformer into a high throughput model, capable of compressing multiple inference inputs into a single input and executing it at a low latency. 

For multiplexing, PruMUX uses the recently proposed   \DataMUX{}~\cite{murahari2023mux}, which appends a multiplexer and demultiplexer as described in Sec~\ref{sec:DataMUX}.  With width $N$, the inference throughput of the Transformer can be improved by a factor of up to $N$, as each multiplexed input takes the same amount of computing resources as performing inference over a single input.

For model compression, PruMUX can use any  method such as network pruning, distillation, or a combination of the two (such as CoFi).  The goal is to substantially reduce the latency of processing an inference.  
For our experiments, PruMUX uses CoFi as the model compression method.

Training a model with PruMUX consists of three phases as shown in Figure \ref{fig:overview}:
\paragraph{Phase 1: Priming the multiplexed model with the token retrieval objective} We first prime the multiplexed transformer model with a token retrieval task. \citet{murahari2022datamux} introduced this "retrieval warm-up" self-supervised objective (shown below) and found it to be critical to improve the performance of multiplexed models. $L$ is the length of each input sentence.
$I$ is the index of the randomly selected sentence from the input batch.
\begin{equation*}
    L_{retrieval}(\textbf{x}^{1:N}) = \sum_{j=1}^{L} -\log P(\textbf{w}_j^I | \textbf{h}_j^I)
\end{equation*}

\paragraph{Phase 2:  Pre-training and fine-tuning multiplexed models}
The multiplexed models from the previous stage are then pre-trained on large-scale text corpora with the masked language modeling (MLM) objective. The pre-trained multiplexed models are then fine-tuned on downstream tasks to yield task-specific multiplexed models.

\paragraph{Phase 3: Model compression} Finally, we use CoFi to jointly prune coarse-grained and fine-grained units in the multiplexed Transformer model. The coarse-grained units include entire attention heads, attention layers, and fully connected layers. The fine-grained units include hidden dimensions and intermediate dimensions of the Transformer model. The demultiplexer's input dimension is pruned in order to match the pruned hidden dimension of the Transformer model. During the pruning process, CoFi uses knowledge distillation to transfer knowledge from the teacher model, i.e., the task-specific multiplexed model, to the pruned model.

\subsection{Implementation Details}

We use the pre-trained multiplexed BERT-base models \cite{murahari2023mux} with the standard BERT pre-training recipe with the masked language modeling objective for $N=2,5,10$ on Wikipedia~\cite{wikidump} and BooksCorpus~\cite{bookscorpus} datasets.
We prime the multiplexed model before pre-training with the token-retrieval task in Section~\ref{sec:DataMUX} on the Wikipedia and BooksCorpus datasets. We then train the pre-trained multiplexed models on the four largest GLUE Tasks~\cite{wang2018glue} -- MNLI~\cite{mnliwilliams2018broad}, QNLI~\cite{wang2018glue}, QQP~\cite{qqp}, and SST-2~\cite{sstsocher2013recursive}. We then use the CoFi structured pruning objective to get pruned multiplexed model on each task dataset. The hyperparameters we use for the training process are shown in Appendix \ref{sec:hyperparam}. We perform a single run to train the model for each setting, i.e., task, multiplexer width $N$, model sparsity $s$,  following the training process.

\subsection{Experiments}

\begin{figure*}[t]
\centering
	\subfloat[MNLI]{%
		\includegraphics[width=0.45\textwidth]{figure/prumux-mnli.png}}
	\subfloat[QNLI]{%
		\includegraphics[width=0.45\textwidth]{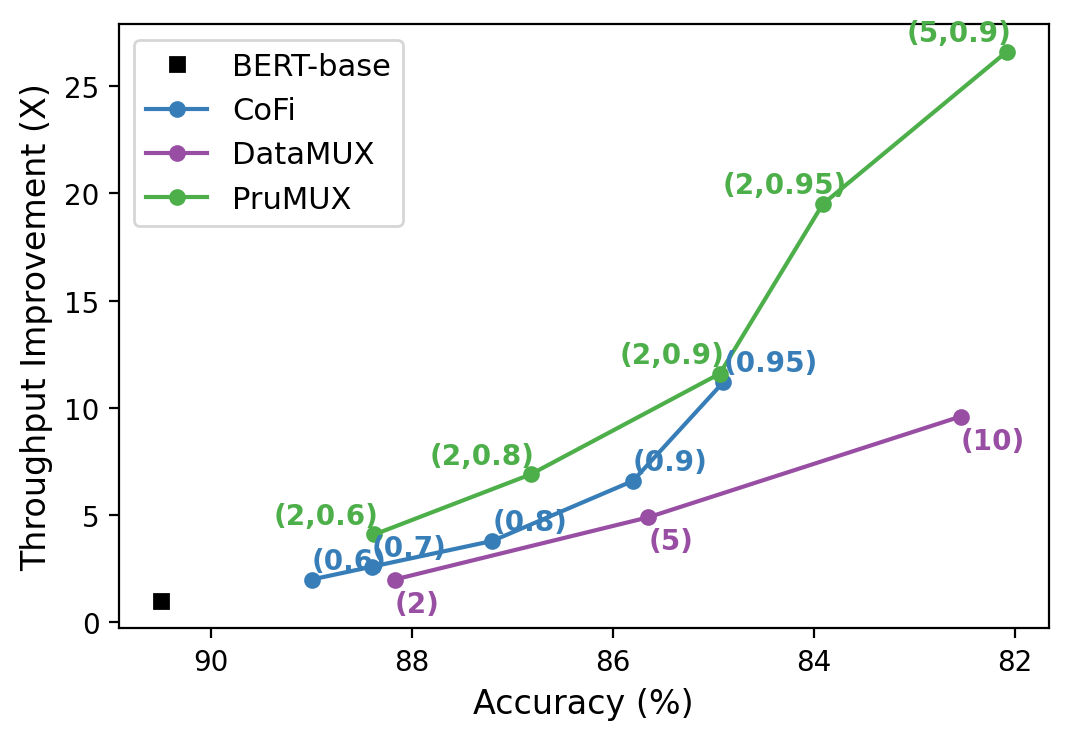}}\\
	\subfloat[QQP]{%
		\includegraphics[width=0.45\textwidth]{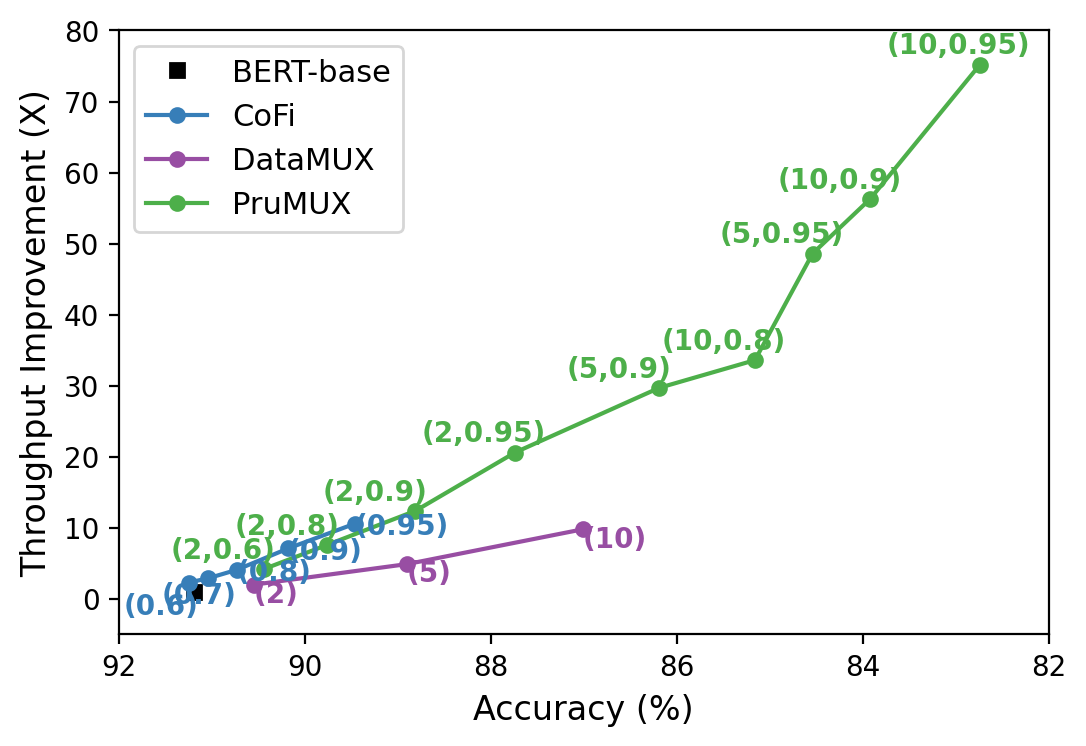}}
	\subfloat[SST-2]{%
		\includegraphics[width=0.45\textwidth]{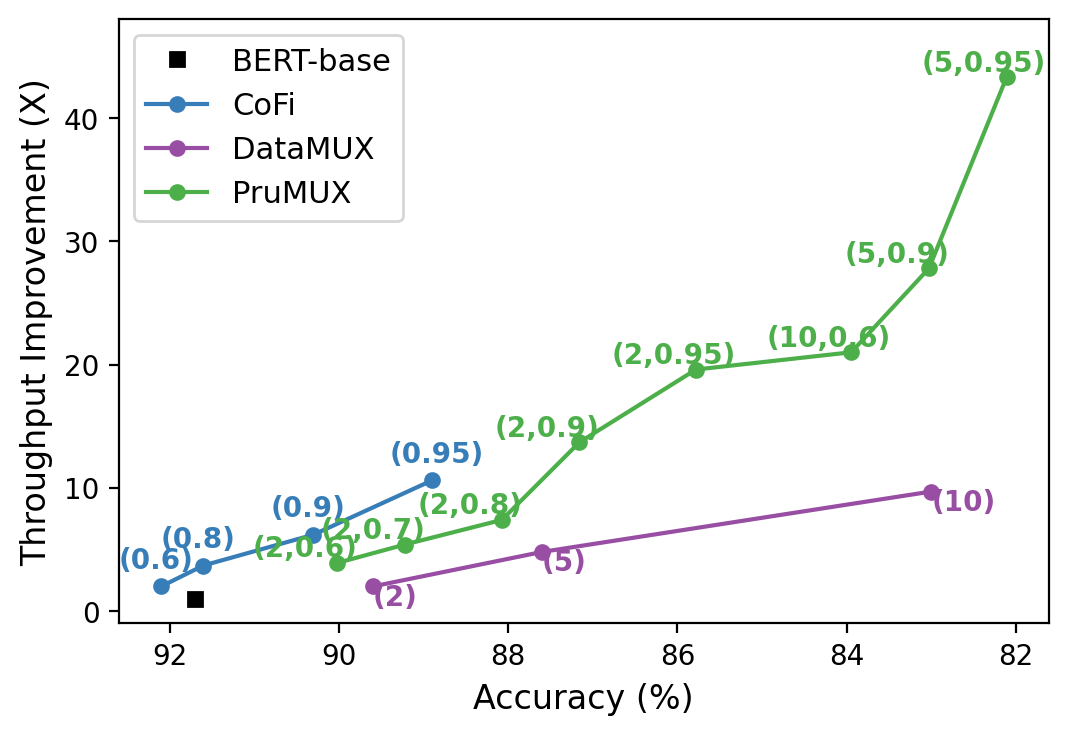}}\\
	\caption[caption for sparse]{\label{fig:overall} Throughput Improvement ($\times$) of PruMUX (ours), \DataMUX{}~\cite{murahari2023mux}, and CoFi pruning~\cite{xia-etal-2022-structured} over the BERT-base model for the MNLI, QNLI, QQP, and SST-2 tasks. The x-axis is the Transformer accuracy, which is inverted to better show throughput improvements of each method for different accuracy loss budgets.
	}
\end{figure*}

\paragraph{Setup}
We would like to answer the question that given an accuracy threshold, whether PruMUX method can achieve a higher throughput than either CoFi or \DataMUX{} alone. 

We compare PruMUXed BERT-base model to three baselines:
\begin{itemize}[leftmargin=*]
    \item {\bf BERT-base}: BERT-base model trained without data multiplexing and model compression.
    \item {\bf CoFi}: BERT-base model pruned by CoFi~\cite{xia-etal-2022-structured} with sparsity\footnote{Sparsity of 0.95 means 95\% of the Transformer model weights are set to zero.} $s =$ 0.50, 0.60, 0.70, 0.80, 0.90, and 0.95.
    \item {\bf \DataMUX{}}: BERT-base model pre-trained by DataMUX~\cite{murahari2023mux} with the multiplexer width $N =$ 2, 5, and 10.

\end{itemize}

We have applied \prumux{} to the BERT-base model with all combinations of $(N,s)$ for all 4 tasks.
We follow the procedure in~\citet{xia-etal-2022-structured} to calculate throughput improvements for PruMUXed Transformers and all three baselines, i.e. BERT-base, \DataMUX{}, and CoFi. The evaluation batch size is 128*$N$, where $N$ is the multiplexer width.

\paragraph{Results}
Figure~\ref{fig:overall} shows the throughput improvements and accuracies of PruMUXed, \DataMUX{}ed, and CoFi-Pruned Transformers over the Transformer base model on the MNLI, QNLI, QQP, and SST-2 tasks with all available parameters. 

The main takeaway is that \prumux{} achieves higher throughput than either CoFi or DataMUX individually in all cases starting at various accuracy thresholds:
\begin{itemize}
    \item 
For  MNLI, with the accuracy thresholds from 80\% to 74\%, PruMUX achieves 7.5-29.5X throughput improvement over the BERT-base model, whereas CoFi improves by 4.0-10.6X and \DataMUX{} by 2.0-4.9X. 

\item
For QNLI, with the accuracy thresholds from 87\% to 82\%, \prumux{} achieves 4.1-26.6X improvement, whereas CoFi improves by 3.8-11.2X and \DataMUX{} by 2.0-9.6X. 

\item
For QQP, with the accuracy thresholds from 89\% to 86\%, PruMUX achieves throughput improvement over BERT-base by 7.6-29.7X, whereas CoFi improves by 10.6X and \DataMUX{} by 2.0-9.8X.

\item
For SST-2, with the accuracy thresholds from 86.5\% to 83\%, PruMUX improves the throughput by 10.1-27.8X, whereas CoFi improves by 10.6X and \DataMUX{} by 4.8-9.7X.  
\end{itemize}

The results also confirm the intuition that PruMUX with
$(N, s)$ incurs an accuracy loss, loosely speaking, close to the sum of the accuracy loss of \DataMUX{} with $N$ and that of CoFi with $s$.  In general, PruMUX can achieve substantial throughput improvement when there is a decent accuracy loss budget. 

\subsection{Discussion}

The results above find top \prumux{} performance with all parameter pairs $(N,s)$, where $N =$ 2, 5, 10 and $s=$ 0.60, 0.70, 0.80, 0.90, and 0.95, for each accuracy loss budget. Searching for top \prumux{} parameters at a finer parameter granularity will require training and testing on all additional parameter pairs.

Exhaustive tests are impractical.
First, for each $N$, pre-training a \DataMUX{} model with multiplexing width $N$ is time-consuming.  Second,
given each pre-trained model with multiplexer width $N$, different sparsities $s$ provide different throughput and accuracy trade-offs. In order to find the sparsity $s$ with the highest throughput given an accuracy budget, one has to train the model for all possible sparsities. The total training time for the sparsities from 0.60 to 0.95 at the granularity of 0.05 for each $N$ takes over six thousand GPU hours on commodity GPUs, for a small original BERT-base model.
A key question is whether one can automatically find a high-throughput ($N,s$) with a small number of \prumux{} experiments.

%% file: prumuxauto.tex
\section{Auto-PruMUX}

To address the question above, we propose Auto-PruMUX, a method to search for top ($N,s$) parameters, to help practitioners balance the performance vs throughput trade-off.  

Our research question is:
    \textit{Suppose we have some experimental data of \prumux{} and the experimental data of \DataMUX{} and CoFi, how can we find and predict the top parameters $(N,s)$ given an accuracy loss budget?}
    
Our approach is to develop performance models for the accuracy and throughput of \prumux{}. 
We first train \prumux{} models for a set of ($N,s$) combinations and measure both the accuracy and the throughput improvement. We then use this data to fit a throughput model and an accuracy model to predict throughput and accuracy respectively given $(N,s)$ parameters. 

We first discuss how we fit the accuracy and throughput models with a set of sparse data points. Given that we are working with a limited set of data points, we opt to use a simple class of interpolation models for modeling \prumux{} accuracy and use an estimation model for modeling throughput. We then outline how we leverage these models to predict top $(N,s)$ parameters, given an accuracy loss budget. We then demonstrate the effectiveness of the \autoprumux{} in predicting the top parameters across a wide range of accuracy loss budgets.

\subsection{Task Accuracy Model}
We use linear interpolation for our task accuarcy model.
\begin{equation*}
\begin{split}
& f_A(N,s) = \\
& \begin{cases}
A_{1,1}(N,s) & \textbf{N}_0 \le N \le \textbf{N}_1, \textbf{s}_0 \le s \le \textbf{s}_1,\\
...\\
A_{i,j}(N,s) & \textbf{N}_{i-1} \le N \le \textbf{N}_i, \textbf{s}_{j-1} \le s \le \textbf{s}_j,\\
...\\
A_{p,q}(N,s) & \textbf{N}_{p-1} \le N \le \textbf{N}_p, \textbf{s}_{q-1} \le s \le \textbf{s}_q\\
\end{cases}
\end{split}
\label{eq:interpolation}
\end{equation*}

Each term is a linear combination of data multiplexer width and model sparsity.

\begin{equation*}
    A_{i,j}(N,s) = \sum_{a=0}^1 \sum_{b=0}^1 k^{(i,j)}_{ab} N^a s^b
\end{equation*}

The model is fitted on the gathered data of model task accuracy at different multiplexer width and sparsity.

\begin{equation*}
\begin{split}
A_{i,j}(\textbf{N}_i,\textbf{s}_j) &= Acc(\textbf{N}_i,\textbf{s}_j) \\
i &= 1, ..., p, j = 1, ..., q\\
\end{split}
\end{equation*}
where $\textbf{N}$ and $\textbf{s}$ are the range of $N$ and $s$ values used to fit the model.

\subsection{Throughput Model}

We collect the throughput values for all $N$ and $s$ on one task ($task_0$) and use the throughput values as the throughput estimations for all tasks.

\begin{equation*}
    f_T(N,s) = Throu_{task_0} (N,s)
\end{equation*}

\subsection{Predicting $(N,s)$}
\label{sec:autoprumux_prediction}

We use our models, $f_A(N,s)$ and $f_T(N,s)$, to model the accuracy and the throughput of \prumux{} with $N > 1$ and $s>0\%$. $Acc(1,s)$ and $Throu(1,s)$ are the measured accuracy and throughput of CoFi-pruned models. $Acc(N,0)$ and $Throu(N,0)$ are the measured accuracy and throughput of \DataMUX{} models. $Acc(1,0)$ and $Throu(1,0)$ are the performance of BERT-base model.
We search for $(N,s)$ parameters that maximize $\zeta_f$ defined below.
\begin{equation}
    \zeta_f(N, s) = Throu(N, s) \cdot g(Acc(N, s))
\end{equation}
\begin{equation*}
  g(x) =
    \begin{cases}
      1
     & x \ge \xi \\
      0 & x < \xi \\
    \end{cases}       
\end{equation*}
Intuitively, $\zeta_f$ tries to tradeoff task performance and throughput, given an accuracy loss budget $\xi$ with the goal of maximizing the throughput. $g(x)$ provides a mechanism for a strict accuracy threshold - i.e. a model that does not meet the minimum required accuracy will have $\zeta_f = 0$.  

\subsection{Experimental Results}
\label{sec:autoprumux_result}

\paragraph{Experimental setting} 
In this section, we show \autoprumux{}'s prediction results by fitting the performance models using a set of parameter space and predicting top parameters on a larger set of parameter space.

We define the set of ($N,s$) parameter space (test set) as follows.

\begin{itemize}
    \item BERT-base model - $(N,s)$ $N=$ 1, $s=$ 0.00
    \item CoFi models - $(N,s) $ $N=$ 1, $\forall s \in$ 0.60, 0.70, 0.80, 0.90, 0.95
    \item \DataMUX{} models - $(N,s) $ $  \forall N \in$ 2,5,10, $s=$ 0.00
    \item PruMUX models - $(N,s) $ $  \forall N \in$ 2,5,10, $\forall s \in $ 0.60, 0.65, 0.70, 0.75, 0.80, 0.85, 0.90, 0.95\footnote{High sparsity doesn’t work for some $N$s and some tasks, i.e., (5, 0.95), (10, 0.90), (10, 0.95) for QNLI, (10, 0.85), (10, 0.90), (10, 0.95) for SST-2. We exclude these points from our training and test set.}
\end{itemize}

We fit the accuracy model with the model accuracies on ($N,s$)  $\forall N \in$ 2,5,10, $\forall s \in$ 0.60, 0.70, 0.80, 0.90, 0.95 (training set). We fit the throughput model with the throughput of one task on all parameter pairs.

Our goal is to evaluate the task accuracy model, the
 throughput model, and parameter prediction performance.
   
\paragraph{Performance Model Accuracy}
To evaluate the accuracy of the task performance models on the training set, we perform leave-one-out cross-validation for each task. We show the fraction $M_A$ of accuracy predictions with error falling within $\Delta \xi = 1.5\%$ from real accuracy in Table \ref{tab:perfmodel}. 
To evaluate the accuracy of the throughput model on the training set, we fit the model using \prumux{}'s performance of the QQP task.
We show the fraction $M_T$ of throughput predictions with error within 20\% of real throughput improvement in Table \ref{tab:perfmodel}. Across different tasks, our accuracy and throughput models are accurate across a broad set of parameter combinations.

\begin{table}[!h]
\begin{center}
\small
\begin{tabular}{c  c  c c  c  c} 
 \toprule
\textbf{Task} & \textbf{$M_A$} & \textbf{$M_T$} & \textbf{Task} & \textbf{$M_A$} & \textbf{$M_T$} \\ [0.5ex] 
 \midrule
 \textbf{MNLI} & 92.3\% & 92.3\% & \textbf{QQP} & 100\% & 100\%  \\ 

 \textbf{QNLI} & 100\% & 91.7\% & \textbf{SST-2} & 100\% & 100\%  \\

 \bottomrule
\end{tabular}
\caption{Accuracy of the task accuracy model ($M_A$) and accuracy of the throughput model ($M_T$) for PruMUX. \label{tab:perfmodel}}
\end{center}
\end{table}

\paragraph{Top Parameter Prediction}

We show \autoprumux{}'s prediction results by fitting the accuracy model on the training set and fitting the throughput model using the throughput of the QQP task, and predicting top parameters on the test set. We show \autoprumux{}'s top parameter predictions for accuracy loss budget 3\% in Table \ref{tab:autopru-pred}. \autoprumux{} predicts the actual best parameter pairs within its top 3 predictions. In Table \ref{tab:predaccu}, we use \autoprumux{} to predict parameters for accuracy loss budgets in 0\%, 0.5\%, ..., 10\% and show the percentage of accuracy loss budgets which \autoprumux{} predicts the actual best parameter in its top 3 predictions. \autoprumux{} is able to predict top parameters in most cases.

\begin{table}[]
\small
\centering
\begin{tabular}{c c c} 
 \hline
Task &   Auto-PruMUX & Actual Best \\ 
  & ($N,s$), Throu. ($\times$)  & ($N,s$), Throu. ($\times$)\\
 \hline
\multirow{3}{*}{\textbf{MNLI}} & Top 1: (2, 0.65), 4.5$\times$ & \multirow{3}{*}{(2, 0.65), 4.7$\times$} \\
& Top 2: (2, 0.60), 4.2$\times$ & \\
& Top 3: (1, 0.80), 4.0$\times$ & \\
 \hline
 \multirow{3}{*}{\textbf{QNLI}} & Top 1: (2, 0.70), 5.0$\times$ & \multirow{3}{*}{(2, 0.65), 4.5$\times$} \\
& Top 2: (2, 0.65), 4.2$\times$ & \\
& Top 3: (2, 0.60), 3.9$\times$ & \\
 \hline
 \multirow{3}{*}{\textbf{QQP}} & Top 1: (2, 0.90), 12.4$\times$ & \multirow{3}{*}{(2, 0.90), 12.4$\times$} \\
& Top 2: (5, 0.65), 10.6$\times$ & \\
& Top 3: (1, 0.95), 10.6$\times$ & \\
 \hline
 \multirow{3}{*}{\textbf{SST-2}} & Top 1: (1, 0.95), 10.6$\times$ & \multirow{3}{*}{(1, 0.95), 10.6$\times$} \\
& Top 2: (1, 0.90), 6.2$\times$ & \\
& Top 3: (2, 0.70), 5.0$\times$ & \\
 \hline

\end{tabular}
\caption{\label{tab:autopru-pred} 
 \autoprumux{} top 3 $(N,s)$ predictions for different tasks with accuracy loss budgets of 3\% along with their predicted throughput improvements. The actual best parameters $(N,s)$ and their throughput improvements are shown in the last column.
 }
\end{table}

\begin{table}[!h]
\begin{center}
\small
\begin{tabular}{c  c  c  c } 
 \toprule
\textbf{Task} & \textbf{Accuracy}  & \textbf{Task} & \textbf{Accuracy}  \\ [0.5ex] 
 \midrule
 \textbf{MNLI} & 81.0\% &  \textbf{QQP} & 100\%   \\ 

 \textbf{QNLI} & 90.5\% &  \textbf{SST-2} & 90.5\%  \\

 \bottomrule
\end{tabular}
\caption{Percentage of accuracy loss budgets in 0\%, 0.5\%, ..., 9.5\%, 10\% which \autoprumux{} predicts the actual best ($N,s$) parameter in its top 3 predictions.  \label{tab:predaccu}}
\end{center}
\end{table}

%% file: related_work.tex
\section{Related Work}

\subsubsection*{Model Compression}

Model compression reduces the number of model parameters with minimal loss in task performance.  A well-studied method is network pruning, which removes unimportant connections from a network with minimal or no accuracy loss~\cite{lecun89,Hanson89,hassibi1993optimal}.  

Unstructured pruning~\cite{han15, han2015deep, zhu2017prune, frankle19, chen2020lottery, huang2021sparse, sanh2020movement} does not impose any constraints on the locations of non-zero weights.  The resulting network can achieve high sparsity but may not run efficiently on common hardware such as GPUs.  

Structured pruning produces structured sparse matrices that can take better advantage of the parallelism in existing hardware, but its sparsity is relatively lower than the unstructured pruning method for the same accuracy loss budget~\cite{yu2017scalpel, narang2017block, wen2017learning, mao2017exploring, wang2019structured, mcdanel2022accelerating}.  Structured pruning has been applied to transformers to improve inference throughput~\cite{fan2019reducing, sajjad2023effect, voita2019analyzing, michel2019sixteen, prasanna2020bert, chen2020earlybert, mccarley2019structured, hou2020dynabert, yao2021mlpruning}.

Distillation compresses a model by transferring knowledge from a large teacher model to a small student model~\cite{hinton2015distilling}.  
General distillation for Transformer models learn from unlabeled corpus \cite{sanh2019distilbert, sun2020mobilebert, wang2020minilm, turc2019well, jiao2019tinybert}. 
Task-specific distillation for Transformer models learns on task-specific data \cite{sun2019patient}.
\cite{jiao2019tinybert} combines the two distillation methods to improve performance.

Pruning with distillation objective have been explored \cite{sanh2020movement, lagunas2021block}.
\cite{xia-etal-2022-structured} proposes structured pruning with distillation objective to reduce the Transformer parameters by up to 95\% and achieve over 10x speedups with small accuracy drops.

\subsubsection*{Multi-input Multi-output Models}

Multi-input Multi-output models concurrently process multiple inputs within one neural network to reduce network over-parameterization. 
\cite{havasi2021training} and \cite{Rame_2021_ICCV} train independent sub-networks and ensemble them into a multi-input multi-output model to obtain better accuracy and uncertainty estimation with inference cost similar to a single network.
\cite{murahari2022datamux} proposes data multiplexing technique to multiplex multiple input sequences into one input sequence to Transformer model, which leads to up to 18x inference speedup. 
\cite{murahari2023mux} develops pre-trained multiplexed language models to improve model throughput.

\subsubsection*{Performance Modeling}

Various methods have been proposed to estimate the performance of machine learning models.
\cite{justus2018predicting} proposes a method to predict CNN execution time for training. They decompose CNN training into several components, estimate the time for each component, and predict the model execution time as the combination of different components.
\cite{qi2017paleo, cai2017neuralpower} predict the performance of deep neural networks based on the neural network models' architecture.
\cite{stamoulis2018hyperpower} proposes predictive models for the power and memory of neural networks executing on GPUs.
Machine-learning-based cost models \cite{chen2018tvm, bouzidi2020performance} have been explored to predict program running time.

Interpolation \cite{davis1975interpolation} is widely used in engineering and science \cite{oliver1990kriging, keys1981cubic, 816070}, where function values at discrete data points are collected in experiments and the function values at the intervals between discrete data points are estimated using interpolation methods.

%% file: conclusion.tex
\section{Conclusion}

We propose \prumux{}, a method to combine model compression and data multiplexing to build high throughput transformers.
Our implementation of \prumux{} makes use of CoFi and DataMUX and we show that it achieves substantial throughput improvement over either CoFi or DataMUX for a large range of accuracy thresholds.  

We conclude that the reason that \prumux{} performs well in certain range of accuracy loss budgets is that CoFi and DataMUX improve the throughput of a model in two different dimensions: reducing the latency of an inference and compressing multiple inferences.
When the accuracy loss budget is large, both methods lead to non-linear drops in model accuracy, \prumux{} can achieve much better performance than either approach because it uses more conservative parameters for CoFi and DataMUX before each reaches its bad trade-off point. 

We also present \autoprumux{}, a meta-model to automatically predict high-performance parameter combinations for a desired accuracy on a task.  We show it is promising in predicting parameters without individual data points and additional training.

%% file: limitations.tex
\section{Limitations}

Our experiments are limited to 3 DataMUXed pre-trained models ($N=$ 2, 5, and 10) due to compute constraints.  More pre-trained models with different $N$'s would provide \prumux{} with more options to improve throughput and would allow us to conduct a more detailed evaluation of Auto-PruMUX.

PruMUX uses CoFi as its model compression method. Experiments with other methods could improve our understanding of the interactions between model compression and data multiplexing.

\section{Acknowledgement}

Karthik Narasimhan and Vishvak Murahari gratefully acknowledge support from the Samsung GRO program.

%% file: appendix.tex
\appendix
\vfill\eject
\section{Appendix}
\label{sec:appendix}

\subsection{Hyperparameters for Model Training}
\label{sec:hyperparam}

Table~\ref{tab:cofihyper} and Table~\ref{tab:prumuxhyper} show the hyperparameters used in CoFi training and PruMUX training correspondingly. The hyperparameters are from CoFi's open-sourced code. \footnote{\href{https://github.com/princeton-nlp/CoFiPruning}{github.com/princeton-nlp/CoFiPruning}}

\begin{table}[ht!]
\begin{center}
\begin{tabular}{c r r} 
 \toprule
\textbf{Hyperparamter} &\textbf{CoFi}   \\ 
\midrule
 distill layer loss alpha & 0.9, 0.7, 0.5  \\ 
 distill ce loss alpha & 0.1, 0.3, 0.5  \\
 layer distill version & 3, 4, 6  \\
 sparsity epsilon & 0.01  \\
 max seq length & 128  \\
 pruning batch size & 32 \\
 finetune batch size & 64  \\
 training epoch & 20  \\
 finetune epoch & 20  \\
 distill temp & 2 \\
 scheduler type & linear  \\
 prepruning finetune epoch & 1  \\
 lagrangian warmup epoch & 2 \\
 pruning learning rate & 2e-5  \\
 finetune learning rate & 1e-5, 2e-5, 3e-5  \\
 
 \bottomrule
\end{tabular}
\caption{\label{tab:cofihyper} Hyperparameters in CoFi training}
\end{center}
\end{table}

\begin{table}[ht!]
\begin{center}
\begin{tabular}{c r r } 
 \toprule
\textbf{Hyperparamter} & \textbf{PruMUX}  \\ 
\midrule
 distill layer loss alpha &  0.9, 0.7, 0.5 \\ 
 distill ce loss alpha &  0.1, 0.3, 0.5 \\
 layer distill version &  3, 4, 6 \\
 sparsity epsilon &  0.01 \\
 max seq length &  128 \\
 pruning batch size &  32*N \\
 finetune batch size &  64 \\
 training epoch &  40 \\
 finetune epoch &  40 \\
 distill temp &  2\\
 scheduler type &  linear \\
 prepruning finetune epoch &  0 \\
 lagrangian warmup epoch &  2\\
 pruning learning rate &  5e-5 \\
\multirow{2}{*}{finetune learning rate} &  1e-5, 2e-5 \\
   &  3e-5, 5e-5 \\
 
 \bottomrule
\end{tabular}
\caption{\label{tab:prumuxhyper} Hyperparameters in PruMUX training}
\end{center}
\end{table}

\subsection{Dataset Statistics}

Table~\ref{tab:gluestats} shows the sizes and metrics of the datasets in our experiments.

\begin{table}[!h]
\begin{center}
\begin{tabular}{c r r } 
 \toprule
\textbf{Task} &\textbf{Train Size} & \textbf{Metric}  \\ 
\midrule
 MNLI & 393k & accuracy  \\ 
 QNLI & 105k & accuracy  \\ 
 QQP & 364k & accuracy  \\ 
 SST-2 & 67k & accuracy  \\ 

 \bottomrule
\end{tabular}
\caption{\label{tab:gluestats} Data statistics of GLUE datasets}
\end{center}
\end{table}

\subsection{Potential Risks}

Multiplexing with model compression may lead to information leakage between different instances, which can potentially raise privacy concerns if used in a public API serving these models.